\documentclass{llncs}

\usepackage[title]{appendix}

\usepackage{subcaption}
\usepackage{algorithm}
\usepackage{algpseudocode}

\usepackage{fixltx2e}		
\usepackage{algorithm}
\usepackage{xifthen}
\usepackage{graphicx}
\graphicspath{{Figures/}}
\usepackage{amsmath}
\usepackage{booktabs}		
\usepackage{multirow} 		
\usepackage{xspace}
\usepackage{xparse}

\usepackage{algpseudocode}

\algnewcommand\algorithmicforeach{\textbf{for each}}
\algdef{S}[FOR]{ForEach}[1]{\algorithmicforeach\ #1\ \algorithmicdo}

\newsavebox\CBox

\NewDocumentCommand\MGPDris{O{}}{%
	MGPD\ifthenelse{\isempty{#1}}%
	{$^{\mbox{ri}}_{t,w}$}%
	{$^{\mbox{ri}}_{#1}$}%
	\xspace%
}

\NewDocumentCommand\dg{m}{\ifmmode #1^\circ\xspace \else $#1^\circ$\xspace \fi}
\NewDocumentCommand\te{m}{\ifmmode \text{#1} \else $\text{#1}$ \fi}
\NewDocumentCommand\tee{m}{\ifmmode \te{\em #1} \else $\te{\em #1}$ \fi}
\NewDocumentCommand\mb{m}{\ifmmode \mathbf{#1} \else $\mathbf{#1}$ \fi}
\NewDocumentCommand\mi{m}{\ifmmode \mathit{#1} \else $\mathit{#1}$ \fi}
\NewDocumentCommand\mr{m}{\ifmmode \mathrm{#1} \else $\mathrm{#1}$ \fi}
\NewDocumentCommand\mir{mm}{\ifmmode \mathit{#1}_\mathrm{#2} \else $\mathit{#1}_\mathrm{#2}$ \fi}
\NewDocumentCommand\mri{mm}{\ifmmode \mathrm{#1}_\mathit{#2} \else $ \mathrm{#1}_\mathit{#2}$ \fi}
\NewDocumentCommand\mrr{mm}{\ifmmode \mathrm{#1}_\mathrm{#2} \else $\mathrm{#1}_\mathrm{#2}$ \fi}
\NewDocumentCommand\mii{mm}{\ifmmode \mathit{#1}_\mathit{#2} \else $\mathit{#1}_\mathit{#2}$ \fi}
\NewDocumentCommand\vl{mm}{#1 $\pm$ #2} 

\NewDocumentCommand\lrr{m}{\ensuremath{\left({#1}\right)}} 

\begin{document}

\author{Benjamin Patrick Evans, Harith Al-Sahaf, Bing Xue, Mengjie Zhang}
\title{Genetic Programming and Gradient Descent: A Memetic Approach to Binary Image Classification}
\institute{Victoria University of Wellington}

\maketitle

\begin{abstract}
Image classification is an essential task in computer vision, which aims to categorise a set of images into different groups based on some visual criteria. Existing methods, such as convolutional neural networks, have been successfully utilised to perform image classification. However, such methods often require human intervention to design a model. Furthermore, such models are difficult to interpret and it is challenging to analyse the patterns of different classes.  This paper presents a hybrid (memetic) approach combining genetic programming (GP) and Gradient-based optimisation for image classification to overcome the limitations mentioned. The performance of the proposed method is compared to a baseline version (without local search) on four binary classification image datasets to provide an insight into the usefulness of local search mechanisms for enhancing the performance of GP.

\end{abstract}


%
%
%
%

\section{Introduction}
Image classification is an important area of computer vision, with a wide range of applications. Existing approaches to image classification tend to suffer from at least one of the following three downfalls: low accuracy, low interpretability, or require human input (for example a manually defined architecture). 

Conventional neural networks (CNNs) are the current state-of-the-art method for image classification and were first introduced in 1998 \cite{lecun1998gradient}, however, saw a resurgence in popularity after the success of AlexNet \cite{krizhevsky2012imagenet} at the 2012 ImageNet Large Scale Visual Recognition Competition \cite{ILSVRC15}. Since the release of AlexNet, every winner of the ImageNet challenge has used a CNN of some sort \cite{ILSVRC15}. 2013 was Clarifai, 2014 was VGGNet \cite{simonyan2014very}, 2015 was ResNet \cite{he2016deep}, 2016 was CUImage and finally 2017 was BDAT. From 2015, CNNs began outperforming human professionals. The key idea behind CNNs is the shared weight architecture, where filter coefficients are learnt through backwards propagation and convolved across an image. Other key developments which made this possible are pooling layers for reducing dimensionality (although these still remain controversial \cite{hinton2011transforming}), and new activation functions, e.g., rectified linear units (ReLU) being the most common to avoid the vanishing gradient problems as networks become deeper. The benefits of CNNs are clear from the high performing results. 

There are, however, some limitations with CNNs. The main limitations are that architectures are often manually crafted, and models typically have low interpretability. Various methods have been proposed to overcome these limitations, such as evolving CNN architectures \cite{desell2017large,suganuma2017genetic}. However, this has a huge computational cost associated. Likewise, steps have been taken to improve the interpretability of CNNs, such as DeconvNets \cite{zeiler2014visualizing} that allow visualisation of intermediary layers, and saliency maps \cite{simonyan2013deep} that visualise the gradient of an output class with respect to an input image. However, the fully connected layers (and thus the feature relations) still remain relatively uninterpretable.

Genetic programming (GP) is another method which can be used for image classification and can help overcome some of the aforementioned limitations. GP is an evolutionary computation (EC) technique that mimics the principles of natural selection and survival of the fittest to automatically search the solution space for a user-defined problem. One key benefit here is the architecture of solutions is determined automatically, removing the need for human intervention.

In our previous work \cite{previous}, we proposed a novel method for binary image classification which utilised principles from both GP and CNNs. We found the results to be competitive to general classification algorithms, however, there was still room for improvement as the method did not outperform CNNs. In this work, we look at the effects of incorporating a local search mechanism into the evolutionary process, similar to the updating scheme used in CNNs (backpropagation with gradient descent) as a method of fine-tuning individuals.

\subsubsection*{\textbf{Goals}}
Specifically, the goal is to assess whether local search can improve the quality of solutions found with evolution alone. The objectives are to:

\begin{itemize}
    \item Integrate local search (as gradient descent) into the evolutionary process as a fine-tuning operation
    \item Investigate whether local search combined with evolution can help to improve the solutions over utilising evolution alone
    \item Analyse a good performing solution to gain an understanding of the performance
\end{itemize} 


 \section{Background}
The general area of combining EC methods with local search is referred to as memetic algorithms, and the idea of combining local search mechanisms specifically with GP has been looked at in \cite{trujillolocal}.

While EC methods such as GP are known as Gradient-free optimisation methods, when available, gradient information can be incorporated as a supplement to the evolutionary process. This has been looked at in \cite{zhang2004genetic,smart2004applying}, where gradient descent is used as a local search mechanism in GP for classification. However, the function set is limited to the four basic arithmetic operators ($+$, $-$, $/$, and $\times$) and the method is not tailored for image classification. It was found local search + GP outperformed standard GP on all datasets trialled.

A different strategy is tested in \cite{emigdio2015local}. While \cite{zhang2004genetic} evolved numeric terminals with gradient descent, \cite{emigdio2015local} assigns a weighted coefficient $\theta$ to each node in a tree. Hence, rather than the output of a node being $n$ the output is now $\theta n$ and $\theta$ is periodically optimised using the Trust Region algorithm. Again, it was shown that in most cases GP with local search significantly outperformed standard GP. It was also noted that the local search did not decrease the interpretability of the evolved models, in fact, resulted in smaller trees overall than standard GP.

The process of Local search within GP is referred to as ``lifetime learning'' in \cite{azad2014simple}, analogous to how continual development happens to an individual in nature. The method (named Chameleon), works on the functions/nodes in the tree. However, rather than associating a weight with each node as in \cite{emigdio2015local}, Chameleon searches for replacement nodes from the function set (nodes with matching input, output types and arity) for each node in the tree (starting at the root node). This works similar to an ``exhaustive mutation'' for a node, where all the allowable variations are trialled. Again, the proposed method (with local search) was found to outperform standard GP on the datasets trialled.

The focus of this work will be to incorporate gradient descent as a supplement to the evolutionary process in our work from \cite{previous} and to analyse how this local search effects the resulting individuals.

\section{New Method}

\subsection{Program Structure}
\begin{figure}
    \centering
    \includegraphics[width=.5\textwidth]{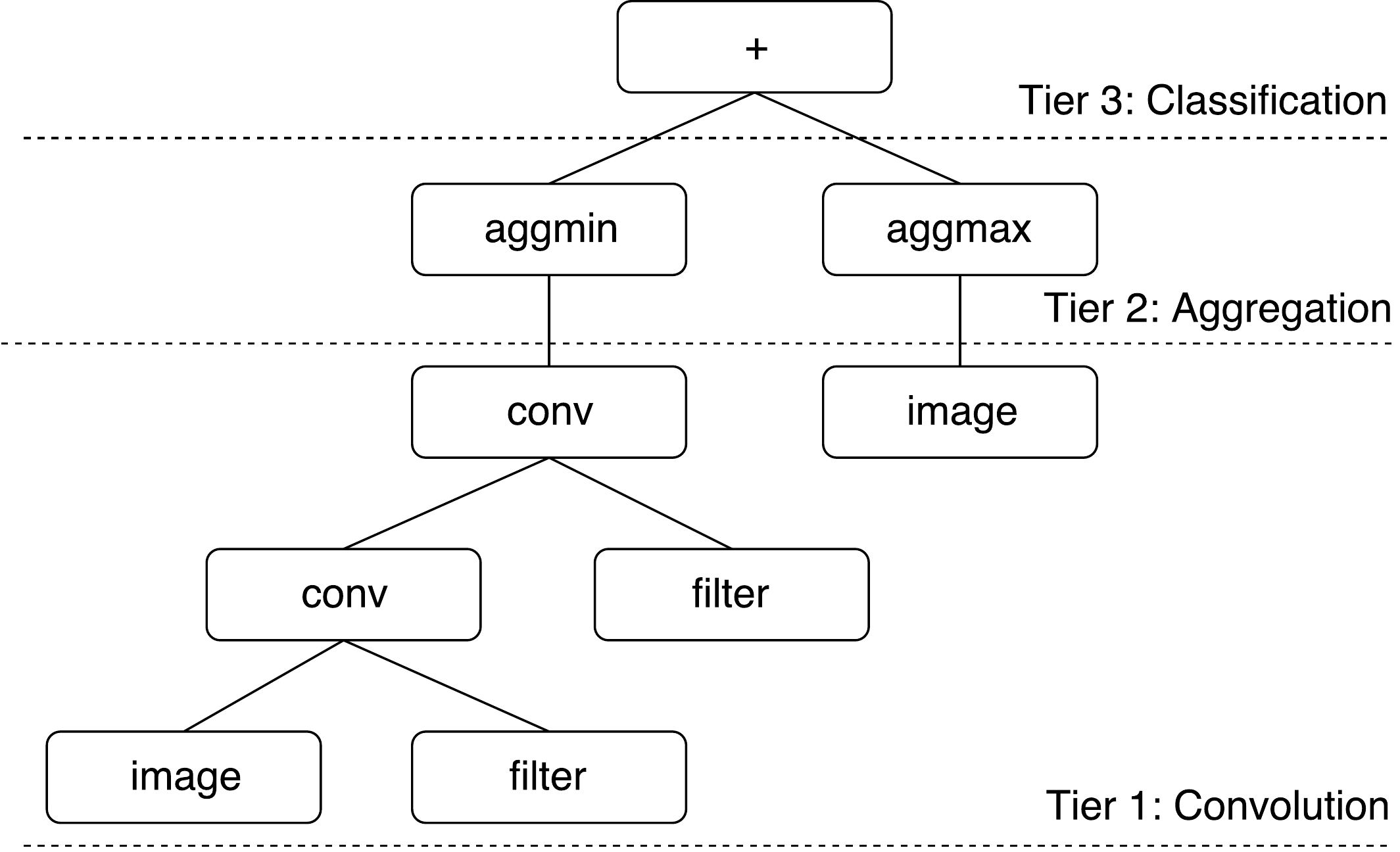}
    \caption{Example tree showing the proposed tiered architecture.}
    \label{figTree}
\end{figure}

While the structure remains the same as proposed in \cite{previous}, it is briefly outlined here for completeness.  There are three tiers in the program structure. Tier 1 (the bottom of the tree) contains the raw image as input and optional convolution and pooling operations. Tier 2 is the aggregation tier, this works over a region of the image and applies a statistical measure such as minimum, maximum, mean and standard deviation, over this region. Tier 3 is a classification tier, which consists of the basic arithmetic operators $+$, $-$, $\times$ and protected $/$ (returns 0 if the denominator is 0). The only compulsory tier is the aggregation tier as this converts an image to a numeric output, whereas tiers 1 and 3 are flexible and can be of different height restricted only by the overall tree size.  An example tree highlighting the structure is given in Figure \ref{figTree}.

Table \ref{tblFunctionSet} summarises the content of the function set, where the input, output and a brief description are provided for each function. The terminals are the raw input image, size and position information for the aggregation window (both specified as a percentage of the image width and height), shape of the aggregation window (either rectangle, column, row or ellipse), a kernel coefficient selected uniformly from the range -1..+1 (to compose the initial values for the 3$\times$3 filters), and an optional random number for the classification tier. A full explanation of these values is given in \cite{previous}.

\begin{table}[t]
    \centering
    \caption{Function set}
    \label{tblFunctionSet}
    {\small
        \begin{tabular}{lllp{8cm}}
            \toprule
            Function & Input & Output & Description \\ \midrule
            $+$ & \multirow{4}{*}{double} & \multirow{4}{*}{double} & \multirow{4}{8cm}{Performs the corresponding arithmetic operator addition, subtraction, multiplication and protected division.} \\
            $-$ & & & \\
            $\times$ & & & \\
            $/$ & & & \\ \midrule
            $\mi{AggMin}\lrr{\cdot}$ & \multirow{4}{2cm}{(image, shape, size, position)} & \multirow{4}{*}{double} & \multirow{4}{8cm}{Applies the statistical measure (minimum, maximum, mean and standard divination) over region (aggregation window) of the image.} \\
            $\mi{AggMax}\lrr{\cdot}$ & & & \\
            $\mi{AggMean}\lrr{\cdot}$ & & & \\
            $\mi{AggStd}\lrr{\cdot}$ & & & \\ \midrule
            \multirow{3}{*}{convolve} & \multirow{3}{2cm}{(image, filter)} & \multirow{3}{*}{image}  & Applies the filter over the input image (uses ``valid'' convolution), and then applies ReLU activation. \\ \midrule
            pool & image & image  & Applies 2$\times$2 max pooling to the input image.\\
            
            \bottomrule
        \end{tabular}
    }
\end{table}

\subsection{Fine Tuning with Gradient-Based Optimisation}

One nice property of the tree structure used is that individuals can be directly represented as nested function calls. For example, a tree can be represented by 
\begin{equation}
    F(x) = \mi{AggMin}\lrr{\mi{conv}(\mi{x}, \mi{filter}), \mi{Rectangle}, 0.1, 0.1, 0.5, 0.5},
\end{equation}
where $x$ (an image) is the only input to the tree/function ($F$), however, if we also consider the filters as parameters to this function, the problem can be represented as 
\begin{equation}
    f\lrr{x, \theta} =  \mathcal{L}\lrr{\mi{targets}, \sigma(F\lrr{x, \theta}},
\end{equation}
where $x$ represents the input images, $\theta$ represents the filter values/coefficients, $\mi{targets}$ represents the true class labels for each $x$, $\sigma$ is the activation function used (in this case sigmoid), and $\mathcal{L}$ is some differentiable loss function. Since the loss function and each function used in the tree are differentiable, this can now be represented as a differentiable optimization problem, where we are trying to find $\theta$ which minimises $f$ by computing  $\frac{\partial f}{\partial \theta}$.


The loss function measures how well we are performing, similar to the fitness function for GP. The classification accuracy was used as a fitness function for GP. However,  classification accuracy is notoriously poor as a loss function for gradient optimization, so we instead use the Cross-Entropy (CE) loss. Mathematically, CE (for binary classification) is defined as
\begin{equation}
    \label{ce}
    \mathcal{L} = - \frac{1}{n}\sum_{i=1}^{n}t_i \log(y_i) + (1-t_i)\log(1-y_i),
\end{equation}
where $i$ indexes the training examples, $n$ is the total number of such examples, $t$ represents the target output, and $y$ the predicted output from the tree. 

This is averaged over each training example as the loss is computed in batches (discussed below), and the size of the batch should not affect the magnitude of the loss.

To compute  $\frac{\partial f}{\partial \theta}$, the chain rule can be applied repeatedly in a manner similar to backwards propagation in neural networks \cite{bprop}. In order to do this, it is necessary that each function be differentiable with respect to the parameters of interest. This is the case for all functions in the tree, and the partial derivatives for each function are given in Appendix \ref{secDerivatives}.

\subsection{Local Search Integration}

To find values of $\theta$ which minimize $f$, gradient descent was used. As the process is run several times throughout the evolutionary process, efficiency is important. Therefore, Stochastic Gradient Descent (SGD) with mini-batches was used in favour of Vanilla Gradient Descent. The algorithm is simple as shown in Pseudocode in Algorithm \ref{algSGD}.

\begin{algorithm}[t]
    \caption{Batch SGD Function}\label{algSGD}
    {\small
        \begin{algorithmic}[1]
            \Function{SGD}{$\mathcal{L}$,$\theta$}
            \While {$\mi{epoch} < 10$}
            \ForEach{$\mi{batch}$}
                \State $\mr{changes}\gets \Call{EvaluateGradient}{\mathcal{L},\mi{batch},\theta}$
                \State $\theta\gets \theta - \lrr{\mi{lr} \times \mi{changes}}$
            \EndFor
            \EndWhile
            \EndFunction
        \end{algorithmic}
    }
\end{algorithm}

Even with mini-batches, the process of gradient descent is relatively slow, as is the overall evolutionary process of GP. Hence, it becomes too large of a computational burden to apply SGD to every generation or to every individual in the population. Three separate methods were trialled, a method without local search (referred to as ConvGP throughout), a method with local search on the 25 fittest individuals every 10th generation starting from the first generation (called ConvGP+LS), and a method (ConvGP+LSE) which does the same as ConvGP+LS, however, on the final generation SGD is run for 100 epochs on the top performer as a ``fine-tuning'' operation (rather than the standard 10 epochs). The reason for this extended training is since this is the final generation the top evolved program will be used as the final model, therefore, the trade-off for extra computational time will likely be worth the improved filter values. 

The evolutionary process works to explore the large search space of potential models (exploration of architectures), the local search is then used to improve upon the top performers found (exploitation of the filter coefficients to minimise $\mathcal{L}$) every 10 generations. ConvGP+LSE exploits this further, by running an extended local search on the final generation.

\section{Experiment Design}

Four widely used image datasets have been used for comparison, from a variety of domains with varying complexities. 

\begin{itemize}
    \item Hands. The hand posture dataset from \cite{triesch1996robust} was used. The classes a (closed fists) and b (open hands) were used. Only images with the "light" background were used.
    \item JAFFE. The Japanese female facial expression dataset from \cite{cheng2010facial} was used. Happy and Sad classes were chosen for comparison. 
    \item Office \cite{gopalan2011domain}. The office dataset features several classes, from several different "domains". For this work, the mobile phone and calculator classes from the webcam domain were used, as the two can look similar making the task of binary classification relatively complex. Images were scaled to 64x64 pixels.
    \item CMU Faces \cite{mitchell1997machine}. For this work, the happy and sad classes were used. Each image also comes in three sizes, only the "small" sized images were used (despite the method working for variable sized images). While featuring similar classes as the JAFFE dataset, the two are quite different in both quality of images and complexity of classification. For example, with the CMU Faces dataset individuals can be wearing sunglasses or facing various directions. 
    
\end{itemize}

To ensure a fair comparison, the exact same training: test splits were used for each of the methods. Three separate seeds were used to shuffle the data, and 30 evolutionary seeds were used for each shuffle (for a total of 90 runs). 

\subsection{Parameters}

Again, to ensure a fair comparison, the parameter settings of the various methods were kept consistent. The values are summarised in Table \ref{tblParameters}, the parameters in \textit{italics} are only relevant for the methods with local search. Note: both the local search methods and the base method had the same number of generations, however, the local search methods run for a longer time due to the inclusion of gradient descent, so training time was not matched.

\begin{table}[]
\centering
    \caption{GP Parameter Values}
    \label{tblParameters}
{\scriptsize
\begin{tabular}{@{}ll|ll|ll@{}}
\toprule
\textbf{Parameter} & \textbf{Value} & \textbf{Parameter} & \textbf{Value} & \textbf{Parameter}                       & \textbf{Value}        \\ \midrule
Population Size    & 1024           & Crossover Rate     & 75\%           & \textit{Epochs}         & 10                    \\
Generations        & 50             & Mutation Rate      & 20\%         & \textit{Number of best} & 25                    \\
Tree Size          & 2 - 10         & Reproduction Rate  & 5\%            & \textit{Learning Rate}  & 0.5                   \\
Tournament Size    & 7              &                    &                & \textit{Batch Size}     & 10\% of training data \\ \bottomrule
\end{tabular}
}
\end{table}

\subsection{Performance Measures}\label{sectionPerformace}

As all datasets used had an equal distribution of images in each class, the raw classification accuracy can be used as a performance measure for evaluation. The classification accuracy formula is given in \ref{eqAccuracy}, this was also used for the fitness function of the program. It is important to note that classification accuracy was not used for the local search, for reasons outlined in Section 3.2.
\begin{equation}
\label{eqAccuracy}
    \footnotesize
    (TP + TN) / (TP + TN + FP + FN)
\end{equation}
Here TP = true positives, FP = false positives, TN = true negatives and FN = false negatives.

\section{Results and Discussions}

\begin{table}[t]
    \caption{Results of the various methods on four datasets.}
    \label{tblResults}
    \centering
    {\scriptsize
        \begin{tabular*}{0.98\textwidth}{@{\extracolsep{\fill}}clllcc}
            \toprule
            
            & & \multicolumn{2}{c}{Accuracy (\%)} & \multicolumn{2}{c}{Time}\\ 
            \cmidrule{3-4} \cmidrule{5-6}
            & Method & \multicolumn{1}{c}{Training} & \multicolumn{1}{c}{Testing} & \multicolumn{1}{c}{Training (m)} & \multicolumn{1}{c}{Testing (ms)}\\
            
            \midrule
            \multirow{3}{*}{\rotatebox[]{90}{Hands}} 
            & ConvGP & \vl{100.0}{0.00} & \vl{95.88}{5.32} & \vl{4.39}{4.11} & \vl{~85.13}{110.24} \\
            & ConvGP+LS & \vl{100.0}{0.00} & \vl{96.39}{5.77} & \vl{5.40}{5.38} & \vl{77.01}{93.68} \\
            & ConvGP+LSE & \vl{100.0}{0.00} & \vl{94.80}{7.00} & \vl{76.38}{95.19} & \vl{108.97}{140.59} \\   
            
            \midrule
            \multirow{3}{*}{\rotatebox[]{90}{JAFFE}} 
            & ConvGP & \vl{99.03}{1.96} & \vl{80.27}{8.29} & \vl{40.92}{47.34} & \vl{142.12}{187.70} \\
            & ConvGP+LS & \vl{98.24}{6.50} & \vl{81.45}{8.04} & \vl{~79.24}{111.47} & \vl{145.70}{216.87} \\
            & ConvGP+LSE & \vl{97.38}{7.51} & \vl{81.77}{9.36} & \vl{159.55}{110.61} & \vl{121.07}{187.81} \\
            
            \midrule
            \multirow{3}{*}{\rotatebox[]{90}{Office}}
            & ConvGP & \vl{96.20}{3.43} & \vl{69.90}{9.46} & \vl{28.55}{22.55} & \vl{66.61}{69.17} \\
            & ConvGP+LS & \vl{92.86}{10.1} & \vl{67.19}{10.6} & \vl{55.92}{26.84} & \vl{38.77}{40.70} \\
            & ConvGP+LSE & \vl{94.77}{9.04} & \vl{67.90}{10.4} & \vl{138.02}{187.10} & \vl{66.92}{50.31} \\
            
            \midrule
            \multirow{3}{*}{\rotatebox[]{90}{Faces}}
            & ConvGP & \vl{69.02}{4.20} & \vl{44.55}{2.91} & \vl{185.04}{194.76} & \vl{393.41}{484.49} \\
            & ConvGP+LS & \vl{68.02}{4.68} & \vl{44.40}{3.28} & \vl{290.15}{286.55} & \vl{259.41}{356.09} \\
            & ConvGP+LSE & \vl{67.81}{3.57} & \vl{45.37}{3.12} & \vl{226.61}{134.79} & \vl{165.64}{205.36} \\
            
            \bottomrule            
        \end{tabular*}
    }
\end{table}

Table \ref{tblResults} summarize the results on the datasets trialled. Results are presented as $\bar{x}\pm s$.   On all four datasets, there was no statistically significant difference in testing accuracy using an unpaired Welch t-test with a significance level of $\alpha = 0.05$. This shows the base ConvGP method is able to evolve good kernel coefficients through evolution alone. The inclusion of local search did not provide any significant improvements in terms of test accuracy, this could be due to one of several main reasons: local search not run for long enough to see improvement, kernel values are in a local optima so local search sees no improvement, potentially overfitting to the training data, or an inappropriate learning rate used. 

\paragraph{Hands} The hands dataset is relatively trivial, as the backgrounds are all plain white, and the two classes used (open and closed) are relatively distinctive. All methods achieved similar results, reaching peak fitness (1.00 = 100\% accuracy) around 5 generations in. Only one round of local search was run (on the first generation), as training stops once an individual reaches the maximum achievable fitness. A small improvement was seen with ConvGP + LS over the base ConvGP, although this was not statistically significant. The increased time taken for the ConvGP + LSE method also makes this a poor candidate for such a simple task, as no benefit was seen from the increased exploitive ability. This shows for simple tasks, there was no benefit to utilising local search in addition to evolution.

\begin{figure}
\centering
\begin{subfigure}{.5\textwidth}
    \centering
    \includegraphics[width=\textwidth]{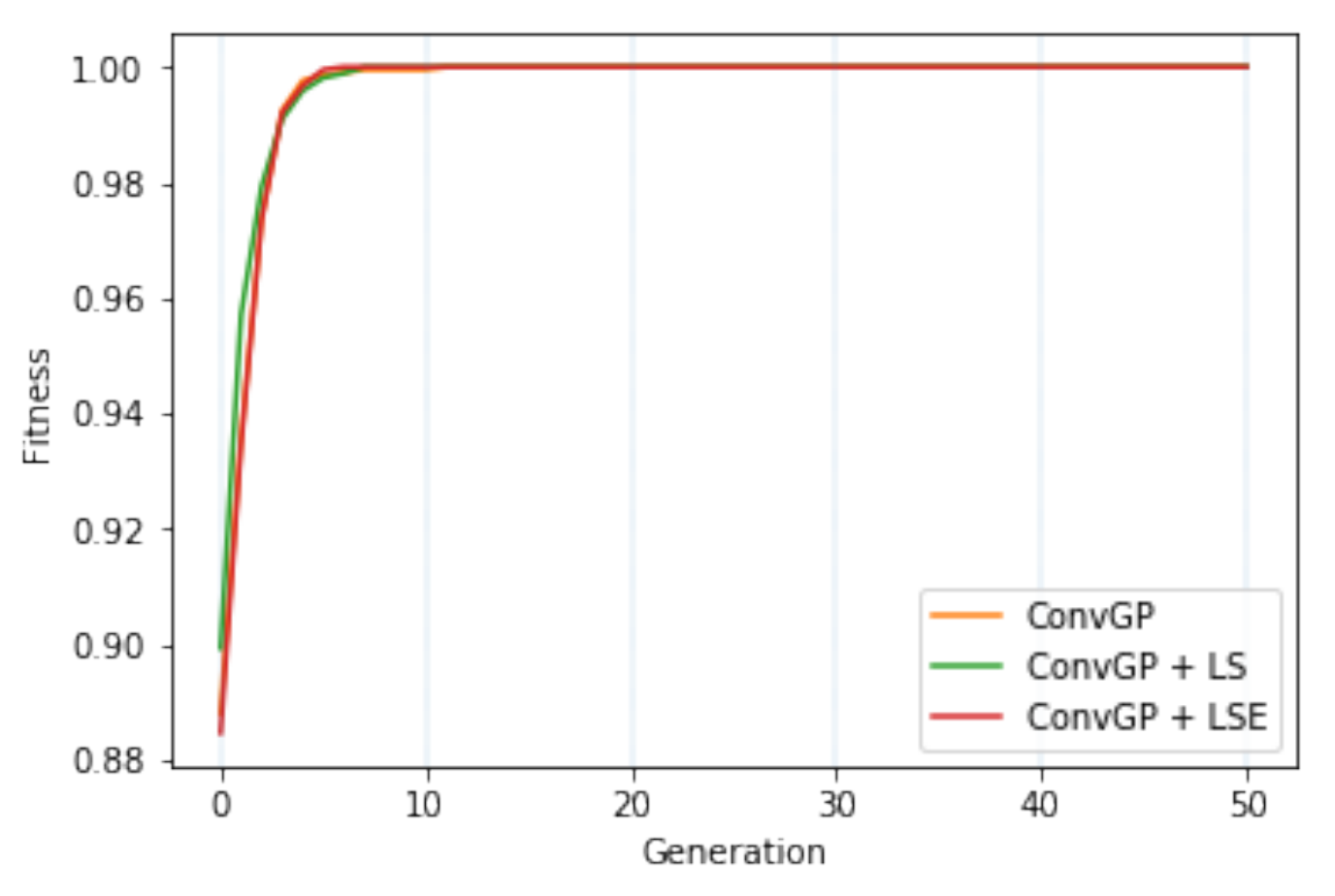}
    \vspace{-1mm}
    \caption{Hands}
    \label{fig:my_label}
\end{subfigure}%
\begin{subfigure}{.5\textwidth}
    \centering
    \includegraphics[width=\textwidth]{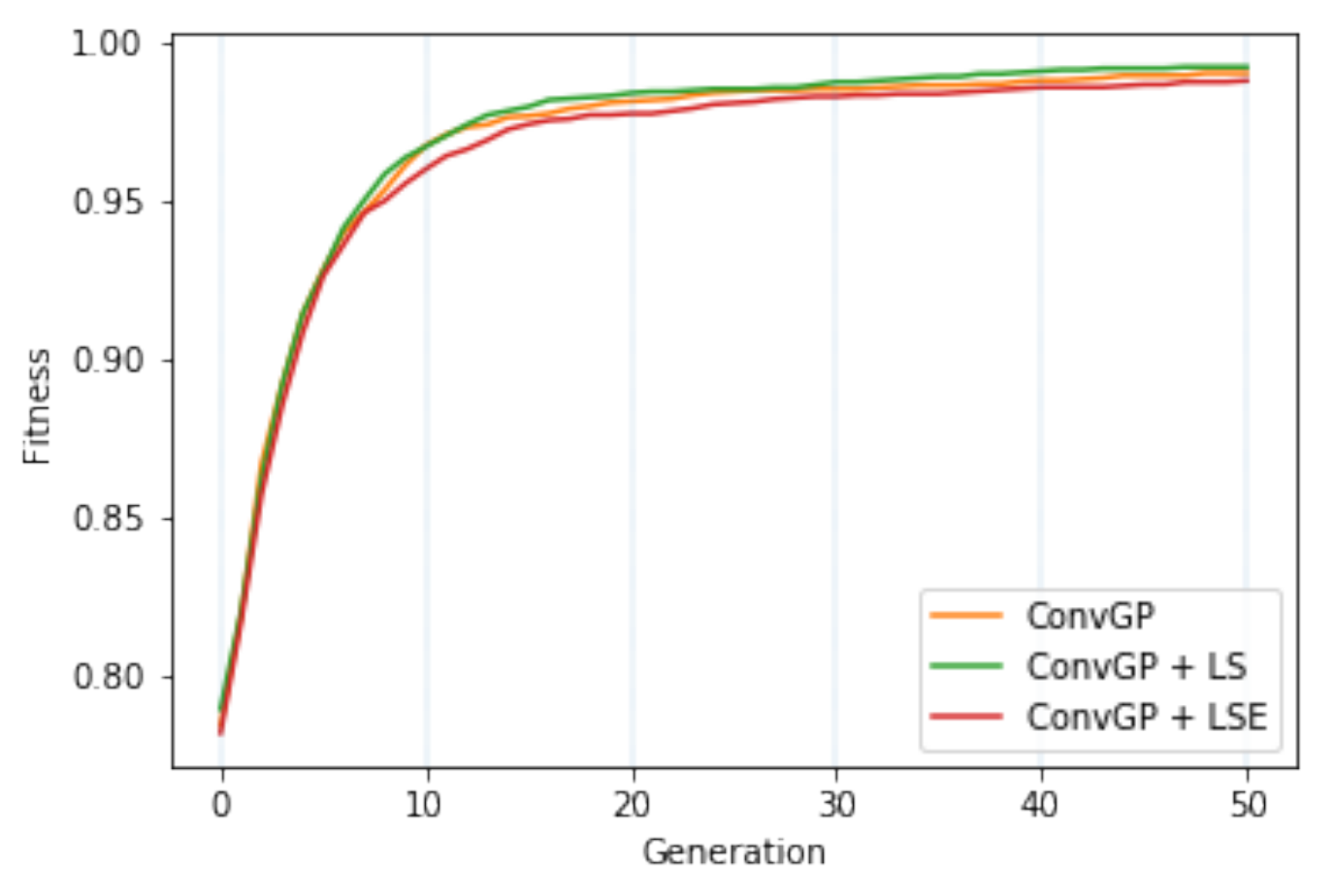}
    \vspace{-1mm}
    \caption{JAFFE}
    \label{figFitness}
\end{subfigure}
\vspace{-2mm}
\caption{Average fittest individual}
\label{fig:test}
\end{figure}

\paragraph{JAFFE} With the JAFFE dataset, a small (but not statistically) significant improvement is seen with the local search methods over the base method. When analyzing the average fittest individuals across the runs, we can see each time gradient descent is run (indicated by the light blue vertical bars every 10 generations) in Figure \ref{figFitness} there is no drastic improvement in subsequent fitness values. This shows with a more exploitative local search (a larger number of epochs), we would likely see additional improvements over the base ConvGP method as currently there is no large impact on the fitness immediately after running.

\paragraph{Office} With the Office dataset, the base ConvGP method actually saw the highest testing accuracy of the tree, although again, this was not statistically significant. While the accuracies were statistically equivalent, the base ConvGP method saw drastically faster training times than the two local search methods. Analysing generational fitness showed similar trends for all three methods, so for this particular dataset local search showed no improvement. 

\paragraph{CMU Faces} Interestingly with the CMU faces dataset, the highest average training accuracy is seen with the baseline ConvGP method. However, for testing accuracy, the ConvGP + LSE method performed the best, although this improvement was not statistically significant (p = 0.09). As expected, the local search methods had slower training time than the base ConvGP method, however, the local search methods resulted in faster testing times. The reasoning for this is looked at in more detail in Section \ref{secFurther}.

\section{Further Analysis}
\label{secFurther}
Although the difference between the two methods was not statistically significant in terms of test accuracy, there were other additional benefits to the method which incorporated local search.

\begin{figure}
\centering
\begin{subfigure}{.5\textwidth}
  \centering
      \includegraphics[width=\textwidth]{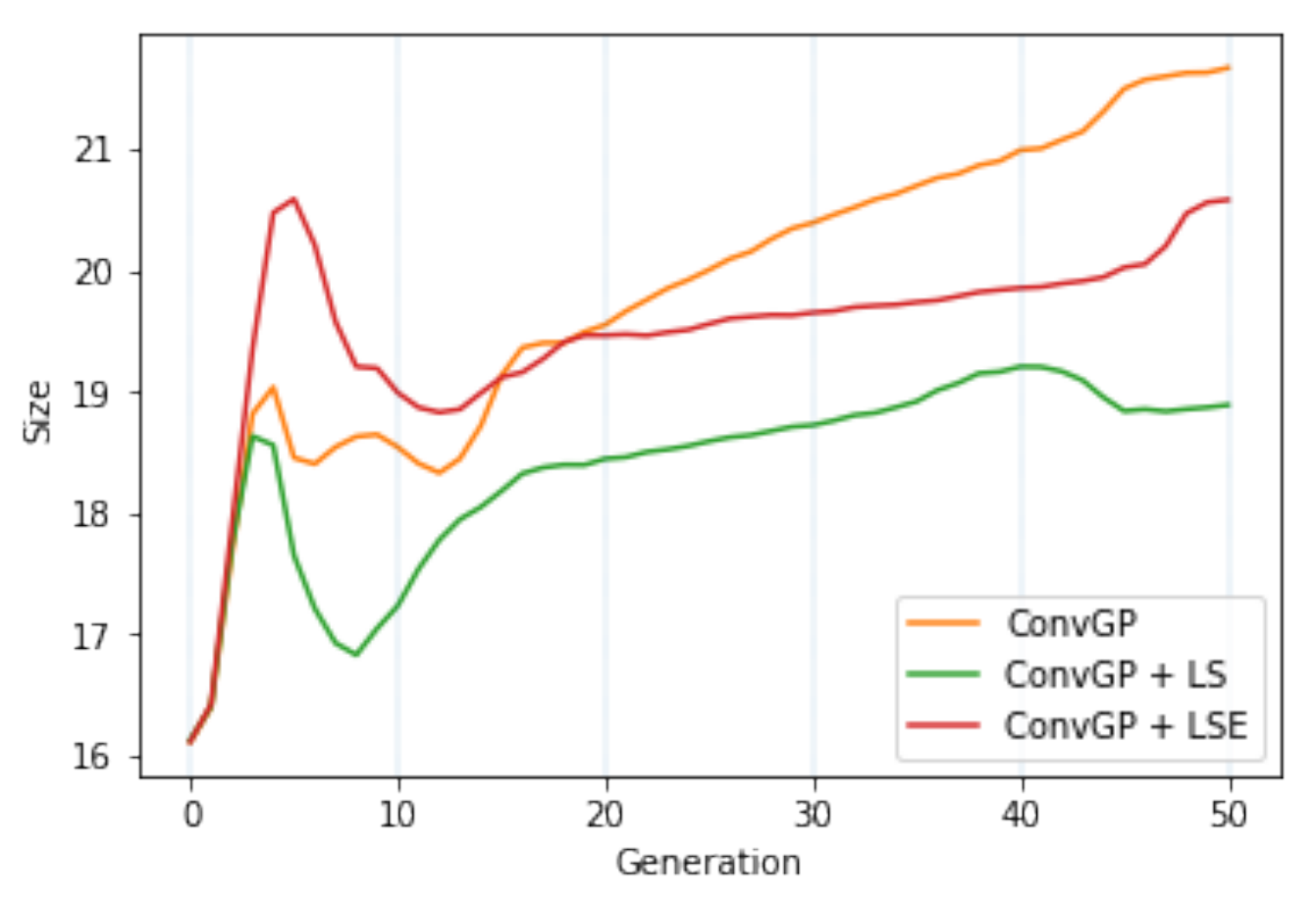}
      \vspace{-3mm}
    \caption{JAFFE}
    \label{figSize}
\end{subfigure}%
\begin{subfigure}{.5\textwidth}
    \centering
    \includegraphics[width=\textwidth]{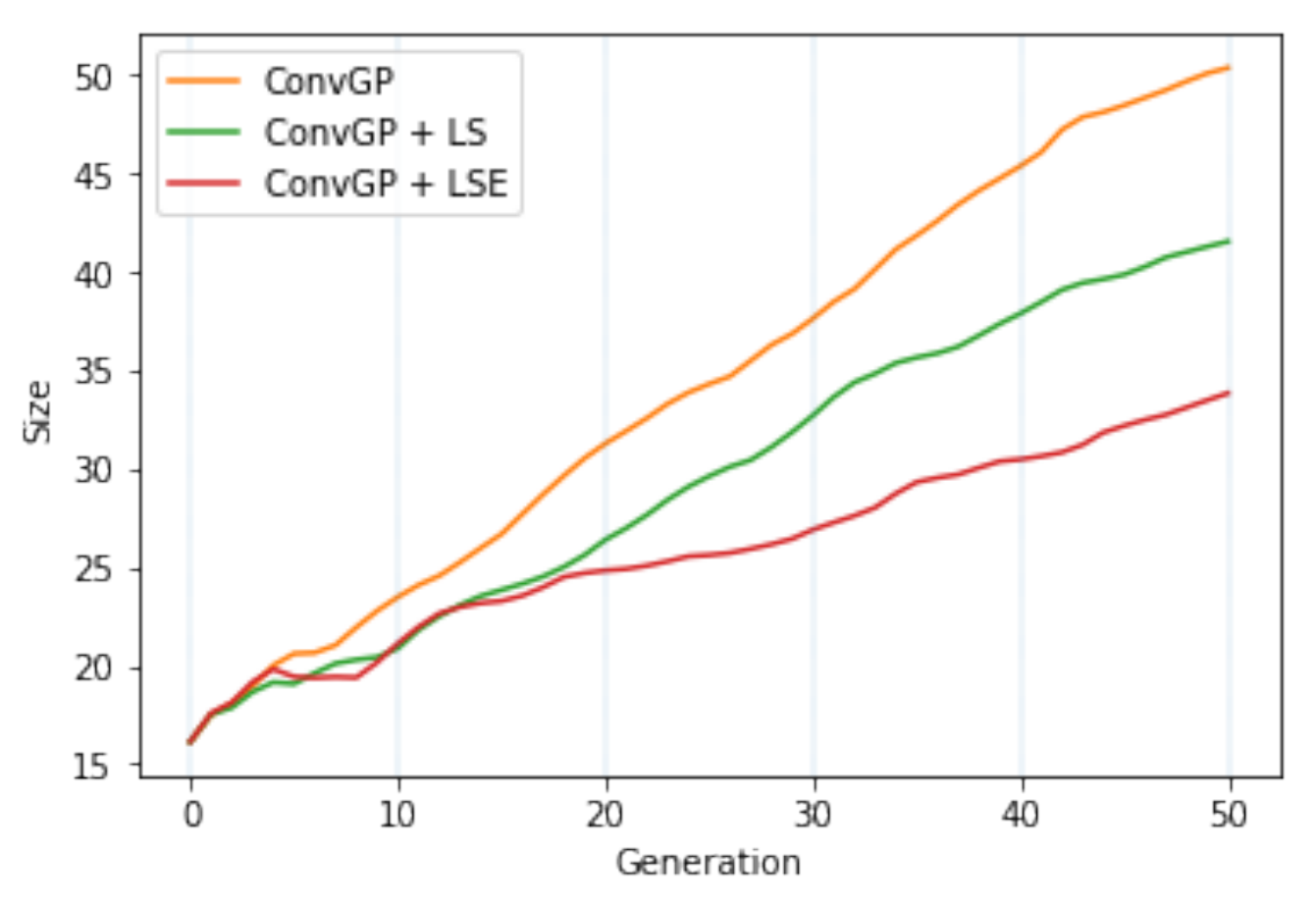}
    \vspace{-3mm}
    \caption{CMU Faces}
    \label{figCMUSize}
\end{subfigure}
\vspace{-2mm}
\caption{Average size of individuals}
\label{fig:test}
\end{figure}

When comparing the average tree size on the JAFFE dataset (Figure \ref{figSize}) and Faces dataset (Figure \ref{figCMUSize}) for each generation across the runs, we can see in both cases the local search methods result in smaller trees on average than ConvGP (a similar trend was seen in \cite{emigdio2015local}). This is likely because we are able to achieve with a single convolution node what would previously require stacked convolutions. It is interesting to note that unlike in \cite{emigdio2015local}, there was no preference for selecting smaller trees when performing gradient descent, so this was an unexpected byproduct of the local search mechanism. This can also be seen when comparing testing time, in many cases, the local search methods result in faster testing times than the base method (at the expense of increased training time).

\begin{figure}
    \centering
    \includegraphics[width=0.5\textwidth]{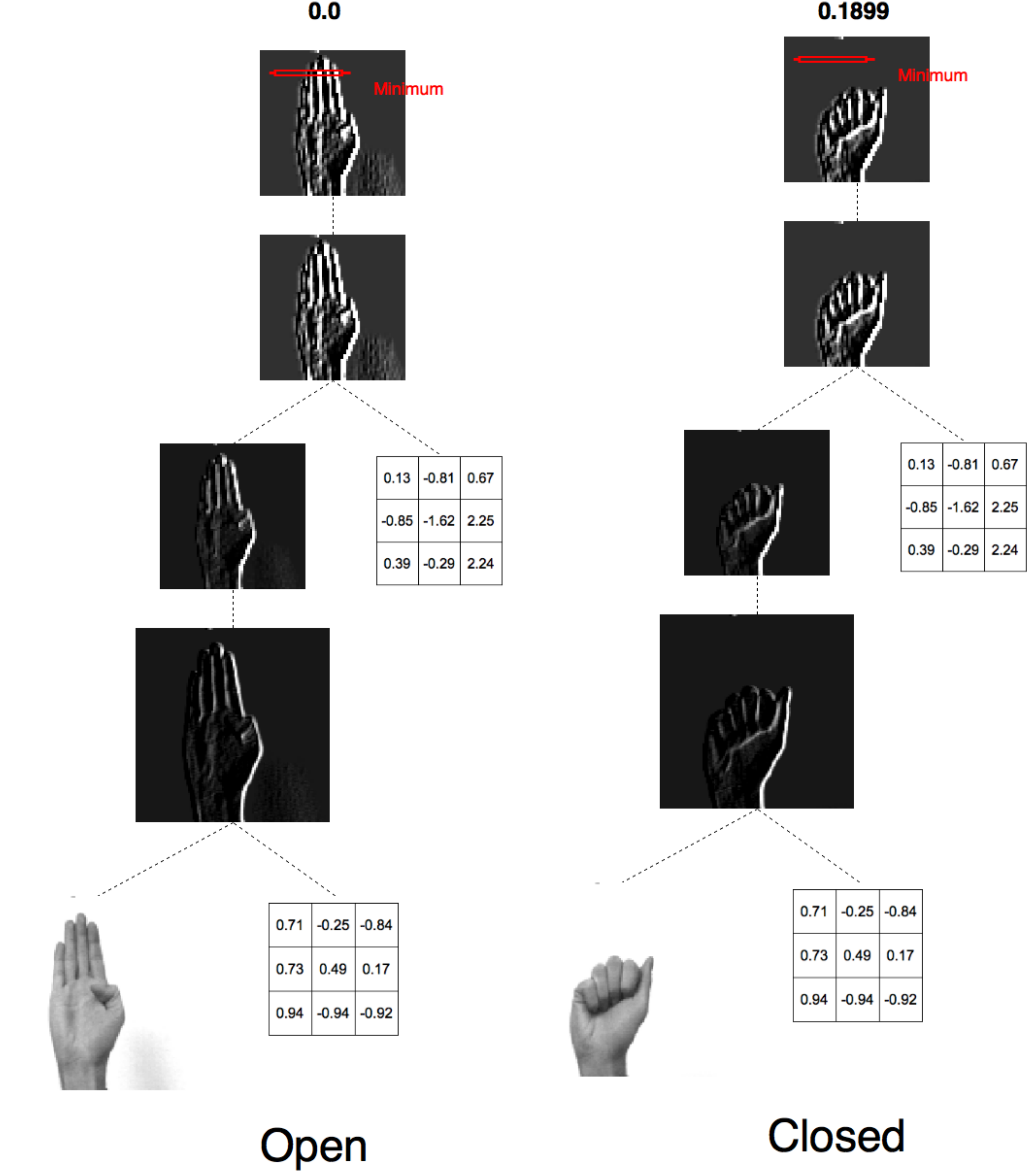}
    \caption{A visualisation of a top performing individual on the Hands dataset}
    \label{figVisual}
\end{figure}

Finally, to show the interpretability (and often simplicity)
of the automatically evolved models, an example is given from the Hands dataset in Figure \ref{figVisual}. This tree was able to achieve 100\% training accuracy. This also shows the flexibility of the proposed structure, while a classification tier is defined, in this instance, no classification tier was required as the output of the aggregation tier was discriminative enough to fully distinguish between the two classes. We can see a key region was identified (where the fingers are present in the open hand images), and high contrast filter values have been learnt. This high contrast, along with the pooling of the image, means fingers show up as completely black (pixel value intensity of 0), whereas the background is a darker grey. The aggregation function used is "minimum", so in the case of fingers in the aggregation window, the value will be 0, and when fingers are not present this will be a grey (in this case 0.1889). The output values are then fed through a sigmoid function, and values $>$ 0.5 are class zero, and values $\leq$ 0.5 are class 1. 

\section{Conclusions and Future Work}
\subsection{Conclusions}
In this work, we looked at the effect of incorporating local search for exploitation of the top performers in a genetic population. It was shown that while the methods which incorporated local search did not see improvement over the raw GP method in terms of classification accuracy, there were other benefits such as reduced tree size on average. With more exploitative local search mechanisms, it is likely an improvement could also be seen in classification accuracy. This shows the benefits of such hybrid (memetic) algorithms for image classification problems. Furthermore, the proposed method was able to achieve the intended goals of achieving high accuracy (as evident by the test result on the datasets trialled), high interpretability as demonstrated in the further analysis section, and the architecture of the solution was able to be evolved automatically rather than manually crafted.

\subsection{Future Work}
As discussed, the local search showed equivalent testing accuracy as the baseline proposed method. With adequate adjustments, local search may be able to significantly improve the baseline method. Namely, the learning rate is currently a manually specified parameter. There has been work done on adaptive learning rates for CNNs, such as Adagrad \cite{duchi2011adaptive} and Adadelta \cite{zeiler2012adadelta}, implementing such methods here would remove the need for a manually specified learning rate and also ideally see improved performance.  Future work could also look to perform a more exhaustive search over the number of epochs for each local search process, as this was computationally prohibitive here due to training time, however, it is likely an increase in epochs would show an improvement in classification performance. Finally, to prevent any over-fitting occurring with an increased number of epochs, a validation set should be introduced to track how the loss is progressing throughout the epochs and stop when this loss no longer improves.

Another current downfall is the training time. This could be remedied in the future, by optimising parts of the program. For example, tree evaluation and local search are both performed sequentially, however, these could, of course, be run in parallel as there is no interaction between the trees. Doing so would result in drastic speedups for the training time. Likewise, no GPU was used for training, an updated program could make use of GPU optimisation for example for the convolutions as shown in \cite{podlozhnyuk2007image}, which again would result in a drastic speedup.

\bibliographystyle{IEEEtran}
\bibliography{Bibliography} 

\clearpage

\begin{subappendices}
\section{Partial Derivatives}
\label{secDerivatives}
The derivatives are given in terms of a single training example for clarity, to avoid subscripts and summations throughout.

As discussed, the output of the tree and the target output was used to compute the loss with the cross-entropy function. Using this information, the partial derivative w.r.t the output of the program can be simplified to
\begin{equation}
    \frac{\partial \mathcal{L}}{\partial x} = y - t,
\end{equation}
where $x$ is the output of the tree (pre activation), $t$ is the target output, and $y$ is the predicted output (post activation with sigmoid). To show this, we can compute the derivative w.r.t to the loss functions input $y$ as
\begin{equation}
\begin{split}
    \frac{\partial \mathcal{L}}{\partial y}
    & = \frac{\partial (-t*log(y) - (1-t)*log(1-y) ) }{\partial y}\\
    & =  \frac{\partial (-t*log(y)) }{\partial y} + \frac{\partial (- (1-t)*log(1-y) ) }{\partial y}\\
    & =  -\frac{t}{y} + \frac{1-t}{1-y}\\
    & =  \frac{y-t}{y(1-y)},
\end{split}
\end{equation}
since the input goes through a sigmoid activation, we can substitute in the derivative for the sigmoid function as
\begin{equation}
    \frac{\partial \sigma}{\partial x} = \sigma(x) * (1 - \sigma(x)),
\end{equation}
and with the chain rule
\begin{equation}
\begin{split}
    \frac{\partial \mathcal{L}}{\partial x}
    &= \frac{\partial  \mathcal{L}}{\partial y} \frac{\partial \sigma}{\partial x} \\
    &=  \frac{\sigma(x)-t}{\sigma(x)(1-\sigma(x))} * \sigma(x) * (1 - \sigma(x)),
\end{split}
\end{equation}
since $y = \sigma(x)$, this simplifies to
\begin{equation}
    \frac{\partial \mathcal{L}}{\partial x} =  y - t,
\end{equation}

The convolution tier has two functions, convolution and pooling. These functions behave in the same manner as when used in CNNs, however, the derivative calculations are often abstracted away with most modern CNN frameworks. Hence, the formulae have been outlined below (a complete description is given in \cite{grad-notes}).

Convolution takes two inputs, an image and a filter. We are interested in the partial derivative w.r.t. each of these inputs. When computing the gradient of a convolution, the result is also a convolution. 
%
Firstly, computing the derivative w.r.t a single value/weight in the filter $w$. If the input image $x$ has height $N_1$+1, and width $N_2$+1 then
\begin{equation}
    \label{wi_1}
    \frac{\partial \mathcal{L}}{\partial w_{a,b}} = \sum_{r=0}^{N_1}\sum_{c=0}^{N_2} \frac{\partial \mathcal{L}}{\partial y_{r,c}} \frac{\partial y_{r,c}}{\partial w_{a,b}},
\end{equation}
where $a$ and $b$ specify a row and a column, $w$ specifies the weight matrix/filter, and $y$ is the result/output. The summation is over the entire image, as each weight in the filter effects every pixel in the image. The first value of the summation is known from the parent node in the tree, but then second value must be calculated as
\begin{equation}
    \frac{\partial y_{r,c}}{\partial w_{a,b}} = x_{r+a, c+b},
\end{equation}
where $x$ is the original/input image. Substituting this back into Equation \ref{wi_1} gives
\begin{equation}
    \frac{\partial \mathcal{L}}{\partial w_{a,b}} = \sum_{r=0}^{N1}\sum_{c=0}^{N2} \frac{\partial \mathcal{L}}{\partial y_{r,c}} x_{r+a, c+b},
\end{equation}
which itself is a convolution, therefore can be more succinctly represented as 
\begin{equation}
    \frac{\partial \mathcal{L}}{\partial w_{a,b}} = \mi{conv}\lrr{x, \frac{\partial \mathcal{L}}{\partial y}},
\end{equation}
Here, we see the result is a convolution between two $N_1\times N_2$ matrices, giving the gradient for an individual weight in the filter.

Now, w.r.t to the input image $x$, assuming the kernel has height $K_1$+1 and width $K_2$+1
\begin{equation}
    \label{xi_1}
    \frac{\partial \mathcal{L}}{\partial x_{r,c}} = \sum_{a=0}^{K_1}\sum_{b=0}^{K_2} \frac{\partial \mathcal{L}}{\partial y_{r-a,c-b}} \frac{\partial y_{r-a,c-b}}{\partial x_{r,c}},
\end{equation}
where the summation is over all the affected pixels, which are the neighbouring pixels within the filters window. Again, the first value of the sum is given from the parent node, but the second value must be computed as
\begin{equation}
    y_{r-a,c-b} = \sum_{a'=0}^{K_1}\sum_{b'=0}^{K_2} x_{r-a+a', c-b+b'} w_{a', b'},
\end{equation}
which simplifies greatly when deriving to
\begin{equation}
    \frac{\partial y_{r-a,c-b}}{\partial x_{r,c}} = w_{a', b'}.
\end{equation}
Hence, now Equation \ref{xi_1} can be rewritten as
\begin{equation}
    \frac{\partial \mathcal{L}}{\partial x_{r,c}} = \sum_{a=0}^{K_1}\sum_{b=0}^{K_2} \frac{\partial \mathcal{L}}{\partial y_{r-a,c-b}} w_{a, b},
\end{equation}
which again can be represented as a convolution, simply by flipping the weight matrix giving
\begin{equation}
    \frac{\partial \mathcal{L}}{\partial x} = conv(\frac{\partial \mathcal{L}}{\partial y}, flip(w)).
\end{equation}
Therefore, the partial derivative for the image and filter are both able to be computed using convolutions.

Pooling is much simpler. Pooling takes a single input (the image), assuming the stride and size are fixed as is the case here. The derivative w.r.t this image is then a matrix of the same shape as the image, where non-maximum pixels have a derivative of zero and max-pixels have a derivative of one (as only max pooling was used in this study). Formally,
\begin{equation}
    \frac{\partial p}{\partial x_i} = 
    \begin{cases}
        1& \text{if } x_i = max(x)\\
        0              & \text{otherwise}
    \end{cases}
\end{equation}
where $p$ is the pooling function, $x_i$ is the $i$th pixel, and $x$ is the pooling window.

The aggregation function takes five parameters, an image, a shape, a $x$ and $y$ position, and a width and height. As we are only interested in updating the filter values, the last four parameters can be safely ignored as the partial derivative w.r.t these terminals will not be used as they represent fixed terminals. The partial derivative w.r.t to the image will be a matrix of the same shape as the input image. All values outside the aggregation window will have a value of 0, with values inside (depending on the aggregation function being used) having a value of 1. However, the convolution operator works over the image as a whole, and we can not control the filter values only for specific regions of the image. This means for efficiency the prior gradient can, in fact, just be passed directly through the aggregation function (treating the gradient for the aggregation functions as 1 since the gradient is being multiplied) when propagating the gradients down the tree, rather than computing the real gradient. 

Remembering the functions used in the classification tier are just the basic arithmetic functions ( $+, -, /, x$) operating on two floating point inputs, the partial derivative w.r.t either of the inputs will be a scalar value. Table \ref{class-deriv} summarizes these partial derivatives.

\begin{table}[t]
    \centering
    \caption{Arithmetic operator partial derivatives}
    \label{class-deriv}
    {\small
    \begin{tabular}{lll}
        \toprule
        Function $f$ & $\frac{\partial f}{\partial a}$ & $\frac{\partial f}{\partial b}$ \\
        \midrule
        $a + b $     & $1  $                  & $1   $                      \\
        $a - b $     & $1$                    & $-1$                        \\
        $a * b$      & $b $                   & $a$                         \\
        $a / b$      & $a ^{-1}$ & $-a b^{-2}$ \\
        \bottomrule
    \end{tabular}
}
\end{table}
\end{subappendices}

\end{document}